\documentclass[letterpaper]{article} 
\usepackage{aaai2026}  
\usepackage{times}  
\usepackage{helvet}  
\usepackage{courier}  
\usepackage[hyphens]{url}  
\usepackage{graphicx} 
\usepackage{natbib}  
\usepackage{caption} 
\usepackage{algorithm}
\usepackage{algorithmic}

\usepackage{amsthm}
\usepackage{amsmath, bm}
\usepackage{amsfonts}

\usepackage{upgreek}
\usepackage{svg}
\usepackage{booktabs}
\usepackage[table]{xcolor}  
\usepackage{multirow}
\usepackage{subfig}
\usepackage{makecell}   
\usepackage{fontawesome5}

\usepackage[dvipsnames,table]{xcolor}
\usepackage{enumitem}
\usepackage{subcaption}
\usepackage[utf8]{inputenc} 
\newcommand{\first}[1]{\cellcolor{Orchid!8}\textcolor{Plum!30!black}{\textbf{#1}}}

\definecolor{HeaderBG}{RGB}{255,255,255}   
\definecolor{goodbg}{RGB}{220,255,220}     
\definecolor{badbg}{RGB}{255,215,215}      
\definecolor{medbg}{RGB}{255,245,205}      

\newcommand{\cgood}{\cellcolor{goodbg}\textbf{O}}
\newcommand{\cbad}{\cellcolor{badbg}\textbf{X}}
\newcommand{\cmed}{\cellcolor{medbg}\Large$\triangle$}

\usepackage{newfloat}
\usepackage{listings}
\DeclareCaptionStyle{ruled}{labelfont=normalfont,labelsep=colon,strut=off} 
\floatstyle{ruled}
\newfloat{listing}{tb}{lst}{}
\floatname{listing}{Listing}
\title{Sensor Calibration Model Balancing Accuracy, Real-time, and Efficiency}
\author {
    Jinyong Yun\textsuperscript{\rm 1},
    Hyungjin Kim\textsuperscript{\rm 1},
    Seokho Ahn\textsuperscript{\rm 1},
    Euijong Lee\textsuperscript{\rm 2},
    Young-Duk Seo\textsuperscript{\rm 1}\thanks{Corresponding author.}
}
\affiliations {
    \textsuperscript{\rm 1}Department of Electrical and Computer Engineering, Inha University, Incheon 22212, South Korea\\
    \textsuperscript{\rm 2} Departments of Computer Science, Chungbuk National University, Chungbuk 28644, South Korea\\
    12191632Y@inha.edu, flslzk@inha.edu, sokho0514@inha.edu, kongjjagae@cbnu.ac.kr, mysid88@inha.ac.kr
}

\begin{document}

\maketitle

\begin{abstract}
Most on-device sensor calibration studies benchmark models only against three macroscopic requirements (i.e., accuracy, real-time, and resource efficiency), thereby hiding deployment bottlenecks such as instantaneous error and worst-case latency. We therefore decompose this triad into eight microscopic requirements and introduce \textsc{\textsc{Scare}} (\textbf{S}ensor \textbf{C}alibration model balancing \textbf{A}ccuracy, \textbf{R}eal-time, and \textbf{E}fficiency), an ultra-compressed transformer that fulfills them all. 
\textsc{SCARE} comprises three core components: (1) Sequence Lens Projector (SLP) that logarithmically compresses time-series data while preserving boundary information across bins, (2) Efficient Bitwise Attention (EBA) module that replaces costly multiplications with bitwise operations via binary hash codes, and (3) Hash optimization strategy that ensures stable training without auxiliary loss terms. Together, these components minimize computational overhead while maintaining high accuracy and compatibility with microcontroller units (MCUs).
Extensive experiments on large-scale air-quality datasets and real microcontroller deployments demonstrate that \textsc{\textsc{Scare}} outperforms existing linear, hybrid, and deep-learning baselines, making \textsc{\textsc{Scare}}, to the best of our knowledge, the first model to meet all eight microscopic requirements simultaneously.
\end{abstract}

%

\section{Introduction}
\label{sec:01_intro}

With the widespread adoption of Internet of Things (IoT) devices, it is now feasible to deploy large-scale sensor networks for environmental monitoring. To keep costs low, these networks often rely on inexpensive sensors, which are prone to nonlinear distortions, noise, and sampling delays \cite{koziel2025efficient}. Consequently, a post-processing calibration step for low-cost sensor data is essential, and, in particular, real-time on-device calibration at the edge is critical. 
On-device sensor calibration upgrades low-cost sensor by mapping raw readings from an inaccurate, low-cost one to values that align with those of a high-quality reference sensor. Traditionally, research on on-device sensor calibration has focused on three requirements: accuracy, real-time performance, and resource efficiency \cite{ahn2025real}. 

However, these macroscopic requirements alone fail to capture the subtle bottlenecks that arise during real-world deployment, such as sampling delays, and hardware (HW) compatibility issues. For example, a model with a low RMSE (Root Mean Squared Error) may still incur large instantaneous errors when environmental conditions change abruptly. \cite{wan2020alert}. To bridge this gap, we decompose the three conventional triad into eight microscopic requirements (Section \ref{sec:02-micro_requirement}). All eight must be satisfied to guarantee trustworthy on-device operation, but existing calibration models typically fulfill only a subset of them \cite{concas2021machinelearningcalibration, villanueva2023mutlisensor, ahn2025real}.

To address these challenges, we propose \textsc{\textsc{Scare}} (\textbf{S}ensor \textbf{C}alibration model balancing \textbf{A}ccuracy, \textbf{R}eal-time, and \textbf{E}fficiency), an ultra‑compressed transformer designed to satisfy all eight requirements on resource-constrained devices such as microcontroller units (MCUs). \textsc{\textsc{Scare}} first applies a bin-level patching scheme that groups tokens into bins and captures salient patterns spanning multiple bins, thereby lowering attention cost without significant information loss (Section \ref{sec:SLP}). It then leverages an efficient bitwise attention mechanism to reduce energy consumption by eliminating redundant operations (Section \ref{sec:LHA}). Finally, \textsc{\textsc{Scare}} adopts an efficient optimization strategy that delivers robust performance at low-cost without complex auxiliary objectives during training (Section \ref{sec:opt}). Collectively, these designs make \textsc{\textsc{Scare}} compatible with practical HW while preserving the essential balance among accuracy, real-time, and resource efficiency.

Our main contributions are as follows:
\begin{itemize}[noitemsep, leftmargin=1em]
    \item \textbf{Discovery}: We decompose the three overarching requirements into eight fine-grained requirements, providing the first comprehensive checklist for on-device calibration.
    \item \textbf{Model}: We present \textsc{Scare}, an ultra-compressed transformer that couples a Sequence Lens Projector with Efficient Bitwise Attention to deliver high accuracy and efficiency on resource-constrained MCUs.
    \item \textbf{Proof}: Experiments on a large-scale air-quality dataset and real embedded-device deployments demonstrate that \textsc{\textsc{Scare}} meets all eight requirements and achieves state-of-the-art calibration performance.
\end{itemize}


\begin{table*}[t]
\centering
\vspace{0.5em}

\resizebox{\textwidth}{!}{%
\setlength\tabcolsep{8pt}
\begin{tabular}{ccccccccccc}
\toprule
\multirow{2}{*}{\textbf{Category}} &
\multirow{2}{*}{\textbf{Model Type}} &
\multirow{2}{*}{\textbf{Model}} &
\multicolumn{2}{c}{\textbf{Accuracy}} &
\multicolumn{3}{c}{\textbf{Real-time}} &
\multicolumn{3}{c}{\textbf{Resource Efficiency}}\\
\cmidrule(l{0.5mm}r{0.5mm}){4-5}
\cmidrule(l{0.5mm}r{0.5mm}){6-8}
\cmidrule(l{0.5mm}r{0.5mm}){9-11}
 &  &  &
Mean & Instant &
Mean & Max & Sampling &
Complexity & Size & Compatibility \\
\cmidrule(l{0.5mm}r{0.5mm}){1-3}
\cmidrule(l{0.5mm}r{0.5mm}){4-5}
\cmidrule(l{0.5mm}r{0.5mm}){6-8}
\cmidrule(l{0.5mm}r{0.5mm}){9-11}

\multirow{3}{*}{\shortstack{Timeseries\\Forecasting}} & Linear & DLinear &
\cgood & \cbad & \cgood & \cgood & \cbad & \cgood & \cgood & \cgood \\
 & Transformer & PatchTST &
\cgood & \cbad & \cmed & \cgood & \cbad & \cgood & \cmed & \cgood \\
 & Hybrid & Mamba &
\cgood & \cbad & \cmed & \cgood & \cbad & \cgood & \cmed & \cgood \\
\cmidrule(l{0.5mm}r{0.5mm}){1-3}
\cmidrule(l{0.5mm}r{0.5mm}){4-5}
\cmidrule(l{0.5mm}r{0.5mm}){6-8}
\cmidrule(l{0.5mm}r{0.5mm}){9-11}

\multirow{1}{*}{Hash-based} & \shortstack{Trainable Hash} & Ecoformer &
\cmed & \cbad & \cgood & \cgood & \cbad & \cgood & \cgood & \cbad \\
\cmidrule(l{0.5mm}r{0.5mm}){1-3}
\cmidrule(l{0.5mm}r{0.5mm}){4-5}
\cmidrule(l{0.5mm}r{0.5mm}){6-8}
\cmidrule(l{0.5mm}r{0.5mm}){9-11}

\multirow{4}{*}{Calibration} & Linear & NLinear &
\cbad & \cbad & \cgood & \cgood & \cgood & \cgood & \cgood & \cgood \\
 & Hybrid & SenDaL &
\cgood & \cgood & \cmed & \cbad & \cgood & \cbad & \cbad & \cmed \\
 & Transformer & TESLA &
\cgood & \cgood & \cmed & \cgood & \cgood & \cgood & \cmed & \cmed \\


& \first{Transformer} & \first{\textsc{\textsc{Scare}}} &
\first{O} & \first{O} & \first{O} & \first{O} & \first{O} & \first{O} & \first{O} & \first{O} \\
\bottomrule

\end{tabular}}
\caption{Comparison of related studies including \textsc{Scare} against the eight microscopic requirements. \textbf{Symbols}: O = Satisfied, \(\triangle\) = Partially satisfied, X = Not satisfied.
\textbf{Accuracy}: Mean / Instant = mean accuracy / instantaneous accuracy.
\textbf{Real-time}: Mean / Max = mean inference latency / maximum inference latency.
\textbf{Resource efficiency}: Complexity, Size, Compatibility = computational complexity, memory size, and hardware compatibility, respectively.
}
\label{tb:ablation_revised}
\end{table*}

\section{Related Work}
\label{sec:02-background}
We identify eight microscopic requirements for on-device sensor calibration and assess the degree to which existing approaches satisfy them, as summarized in Table \ref{tb:ablation_revised}.

\subsection{Microscopic Requirements}
\label{sec:02-micro_requirement}

Most on-device sensor-calibration studies benchmark their models using only three broad criteria: accuracy, real-time performance, and resource efficiency \cite{rahman2020systematic, ahn24sendal, ahn2025real}. This high-level triad obscures important challenges during deployment. It hides fine-grained bottlenecks such as instantaneous error spikes and worst-case latency, and ignores the strict memory and energy budgets of MCUs. Consequently, models that look competitive on paper can still underperform in real-world deployments \cite{ibrahim2024holistic}.

To close these gaps, we define eight microscopic requirements and conduct an in‑depth analysis of operational bottlenecks across three representative model classes: time-series forecasting, hash-based, and sensor calibration models. For every requirement, Table \ref{tb:ablation_revised} assigns one of three ratings: `Satisfied (O)' when the criterion is explicitly built into the design and empirically verified; `Partially satisfied (\(\triangle\))' when it is considered but either lacks rigorous experimentation or underperforms in validation; and `Not satisfied (X)' when it is entirely omitted.

The eight microscopic requirements map onto three macroscopic dimensions (i.e., accuracy, real-time performance, and resource efficiency) as follows:

\begin{itemize}[noitemsep, leftmargin=1em]
\item \textbf{Accuracy}: Prior works \cite{concas2021machinelearningcalibration, ahn24sendal} usually report only mean error.
A rigorous evaluation must also include instantaneous error, which is the model’s responsiveness to abrupt changes, because a model optimized solely for mean error can still miss local spikes and incur large transient mistakes.

\item \textbf{Real-time}: A single latency figure is insufficient to evaluate real-time performance. Even when the mean inference latency may be low, it can suddenly jump at certain times, raising the maximum inference latency. This can make the sampling interval unstable or excessively long, breaking the timing between data points \cite{yang2021deeprt, zhao2025slopt}. Therefore, comprehensive assessment requires mean inference delay, maximum inference delay, and sampling interval delay.


\item \textbf{Resource efficiency}: This dimension decomposes into computational complexity, memory footprint, and HW compatibility. High complexity quickly exhausts the limited CPU cycles and battery budget on edge devices \cite{ni2025energy, liu2022ecoformer}, and oversized models overflow on‑chip memory, forcing slower off‑chip accesses \cite{lin2020mcunet}. Finally, mismatched HW compatibility leaves unsupported operators or memory layouts that stall or fail on the MCU \cite{liang2023mcuformer}.


\end{itemize}

\subsection{Time-series Forecasting Models}


Time-series forecasting models are usually optimized solely to minimize average error, yet this narrow objective prevents them from achieving instantaneous accuracy. Moreover, nonlinear operations, especially self-attention, cause inference latency to spike as input sequences lengthen, destabilizing the sampling interval and preventing the models from meeting real-time sampling requirements. 

DLinear \cite{zeng2023dlinear} lowers computational complexity with purely linear predictors, but its inability to model nonlinear patterns leads to large instantaneous errors. PatchTST \cite{nie2023timeseriesworth64} reduces sequence length by grouping tokens, but still incurs quadratic self-attention cost and cannot guarantee a stable sampling interval under resource constraints. Mamba \cite{dao2024transformers_ssm} employs a selective state-space structure with time-varying matrices to capture long-term patterns efficiently, but it requires specially optimized kernels when parallel processing units (\emph{e.g.,} GPUs) are unavailable and often suffers from high memory consumption and computational latency on resource-constrained devices. \\
\noindent \textbf{Our contributions.}
\textsc{\textsc{Scare}} integrates a patching-based sequence compression with bitwise attention to ensure low computational complexity while maintaining structural conciseness that guarantees high accuracy with minimal layers, effectively addressing the sampling interval delay.

\subsection{Hash-based Models}
Hash-based models \cite{kitaev2020reformer, liu2022ecoformer} reduce the computational cost of attention by grouping tokens or approximating similarities with hash codes, but they inherently risk degrading accuracy. Hash functions that cannot adapt quickly to input distribution shifts are vulnerable to sudden changes, and sampling delays can force the model to generate hash keys from outdated inputs, causing transient prediction errors to increase sharply. Moreover, most hash-based models have yet to be evaluated on resource-constrained devices such as MCUs. 

EcoFormer \cite{liu2022ecoformer} projects queries and keys into a lower-dimensional space via hashing, achieving linear-complexity attention and notable energy savings. However, the representational capacity diminishes on short sequences, leading to significant performance degradation. In addition, the model depends on specialized GPU kernels for its auxiliary operations, which limits practical adoption on resource-constrained devices. \\
\noindent \textbf{Our contributions.}
\textsc{Scare} uses a sequence-wide patching strategy that preserves accuracy while remaining compatible with resource-constrained HW. In addition, its dynamically updated bitwise attention addresses sampling interval delay and adapts robustly to sudden distribution shifts.

\subsection{Sensor Calibration Models}
Sensor calibration models have largely focused on macroscopic requirements for practical IoT systems. NLinear \cite{zeng2023dlinear,ahn24sendal} normalizes each time-series to its final value, yet its strictly linear design incurs large errors during nonlinear fluctuations. SenDaL \cite{ahn24sendal} blends linear and deep-nonlinear components to reduce mean error, but the added depth inflates worst-case latency. TESLA \cite{ahn2025real} lowers both mean and instantaneous error through logarithmic patching and an efficient attention mechanism, but its memory layout conflicts with MCU's Direct Memory Access (DMA) functions, undermining HW compatibility. Consequently, existing approaches continue to face fundamental trade-offs among accuracy, real-time performance, and resource efficiency.\\
\noindent \textbf{Our contributions.}
\textsc{\textsc{Scare}} satisfies all eight microscopic requirements by reducing both mean and instantaneous errors through a novel patching strategy and by refining its attention mechanism to guarantee real-time performance and HW compatibility on resource-constrained devices.

\section{Efficient On-Device Sensor Calibration Model: \textsc{Scare}}
\label{sec:model}

This section first defines the sensor calibration problem and introduces \textsc{\textsc{Scare}} (\textbf{S}ensor \textbf{C}alibration model balancing \textbf{A}ccuracy, \textbf{R}eal-time, and \textbf{E}fficiency), a method designed to satisfy eight core requirements across three key challenges.

\subsection{Definition of Sensor Calibration Problem}
Sensor calibration is the process of comparing low-cost sensor outputs with reference sensor measurements, learning corrections to reduce distortion and bias, and improve accuracy. \cite{yu2020airnet, ahn24sendal, ahn2025real}



The calibration model is trained by comparing the low-cost sensor’s input vectors \(\mathbf{z}_j\) against simultaneous measurements \(r_j\) from a high-precision reference sensor \(Y\).  
Defining a neural network \(f(\,\cdot\,;\theta)\) with parameters \(\theta\), we solve:

\begin{equation}
  \theta^{\ast} \;=\;\underset{\theta \in \Theta}{\arg\min}\;
  \sum_{j\in\mathcal{T}} \mathcal{J}\bigl(r_{j},\,f(\mathbf{z}_{j};\theta)\bigr)
\end{equation}
where \(\mathcal{J}\) is an objective function.


Reference sensors can reduce bias due to distribution mismatch, and eight additional microscopic indicators of accuracy, real-time, and resource efficiency must be considered when deploying models to embedded devices.


\subsection{Overview of \textsc{\textsc{Scare}}}

\begin{figure}
    \centering
    \includegraphics[width=1\columnwidth]{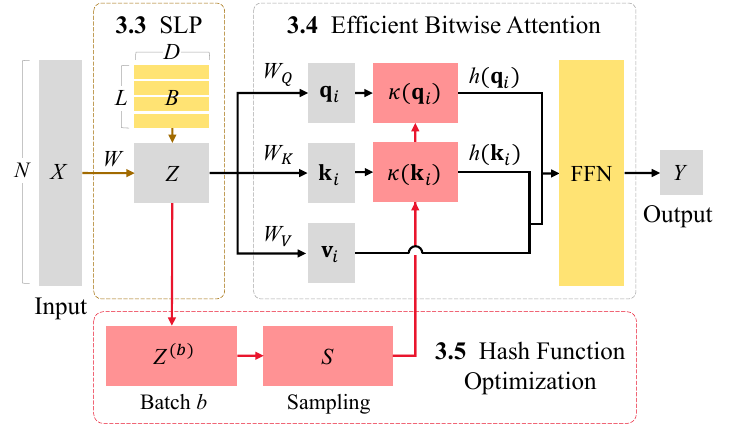}
    \caption{Overview of \textsc{Scare}, composing (i) Sequence Lens Projector (SLP), (ii) Efficient Bitwise Attention (EBA), and (iii) Hash Function Optimization.}
    \label{fig:Scare}
\end{figure}

Figure \ref{fig:Scare} illustrates the overall structure of \textsc{\textsc{Scare}} (\textbf{S}ensor \textbf{C}alibration model balancing \textbf{A}ccuracy, \textbf{R}eal-time, and \textbf{E}fficiency) for on-device sensor calibration in IoT systems. \textsc{\textsc{Scare}} is an ultra-compressed transformer designed to satisfy eight microscopic requirements across accuracy, real-time, and resource efficiency. 

\textsc{Scare} constructs dual embeddings using local and global weight matrices, following the approach of \citealp{ahn2025real}, to capture both instantaneous sensor variations and long-term trends while reducing the input complexity. The input is then processed through three core components.
First, we propose the Sequence Lens Projector (SLP) (Section \ref{sec:SLP}) to compress entire sequences, which compresses sequences across multiple focal distances to reduce computational complexity.
Next, we introduce Efficient Bitwise Attention (EBA) (Section \ref{sec:LHA}), which enables efficient computation through binarized hash codes. EBA replaces expensive multiplications with efficient additions, significantly reducing energy consumption for these resource-constrained environments. 
Finally, we employ effective optimization strategies (Section \ref{sec:opt}) through dynamic sampling and kernel-based approximation without complex auxiliary objectives. These components work as integrated modules, making \textsc{\textsc{Scare}} applicable to practical sensor calibration while satisfying all eight microscopic requirements.

\subsection{Sequence Lens Projector}
\label{sec:SLP}

Applying attention directly to long time-series data incurs a computational cost of $\mathcal{O}(N^{2})$. To mitigate this, various patching methods have been proposed to compress sequence‑level information to improve both accuracy and efficiency \cite{chen2025sequence, nie2023timeseriesworth64}. Most studies rely on hard pooling, which divides each segment into fixed‑size bins and calculates representative values using averages or weighted sums \cite{qiu2025enhancing}.

This approach has fundamental limitations. First, segment‑wise processing makes it difficult to capture sudden changes across bin boundaries. Second, the process of batch‑wise synthesis across the entire sequence incurs imbalanced information levels between bins, causing important patterns to be averaged out and lost. 
Recently, TESLA's proposed logarithmic patching enables flexible pooling, but the non‑uniform patch lengths hinder the implementation of optimized computational structures.
Moreover, its incompatibility with DMA and vector operations leads to inference delays on practical HW devices such as MCUs.

To address these issues, we propose Sequence Lens Projector (SLP). The core idea is that while existing methods are similar to observing the sequence through multiple small lenses with limited scope, SLP observes the entire sequence through a single large lens with multiple focal distances simultaneously. 
By assigning global bias to each bin, SLP ensures that bins reflect global context rather than relying solely on local features, allowing significant patterns to contribute across multiple bins. 
This design reduces boundary loss by preserving patterns that span across adjacent bins, and alleviates inter-bin imbalance by preventing certain bins from becoming over- or under-represented.
Additionally, SLP utilizes a single matrix structure to optimize all bins jointly, achieving efficient information compression without additional parameters. 

Specifically, SLP operates as follows:
\begin{equation}
Z=\mathrm{SLP}(X) = \bigl[W^{\top} X + B\bigr]\in \mathbb{R}^{L \times D},
\end{equation}
where $X \in \mathbb{R}^{N \times D}$ is the input, $W \in \mathbb{R}^{N \times L}$ is the compression weight matrix, and $L = \lceil \log_{2} N \rceil$ is the number of lenses. Each lens is represented by a column vector of $W$, indicating the importance of each time step. The bias matrix $B \in \mathbb{R}^{L \times D}$ provides distinct offsets for each feature dimension and lens, improving accuracy.

During training, SLP automatically optimizes $W$ and $b$ through gradients of the calibration loss. This reduces mean error and prevents sudden spikes in instantaneous error during abrupt signal changes, ensuring stable performance. Consequently, SLP improves both accuracy and efficiency while ensuring compatibility and minimizing inference delays even in resource‑constrained environments such as MCUs.

\subsection{Efficient Bitwise Attention}
\label{sec:LHA}

In sensor calibration, nonlinear models typically leverage standard multi-head self-attention (MHSA) due to their compressed tokens through patching \cite{ahn2025real}. However, this approach may incur computational overhead in resource-constrained environments such as on-device deployment scenarios. In this regard, we propose Efficient Bitwise Attention (EBA) that employs hash attention to optimize energy and memory efficiency and reduces unnecessary operations practical sensor calibration.

Hash attention reduces the cost of self‑attention by converting tokens into hash codes so that only similar tokens interact \cite{kitaev2020reformer, liu2022ecoformer}. 
We first leverage the sign function \cite{rastegari2016xnor, liu2018bi} and straight‑through estimator (STE) to replace multiplications with bitwise operations, reducing computational cost and energy consumption. And we employ the radial basis function (RBF) kernel \cite{liu2022ecoformer} during training to improve the precision of hash approximation. Although the RBF kernel requires additional computation, the added cost is limited due to the input sequence compressed by SLP. 






Formally, given samples \(S=\left[\mathbf{s}_1, \cdots,  \mathbf{s}_m\right]\in\mathbb{R}^{m\times D}\) (as detailed in Section \ref{sec:opt}) with projection matrix \(A \in \mathbb{R}^{m \times c}\), the hash function \(h:\mathbb{R}^{D} \rightarrow \left\{-1, +1\right\}^c\) used to binarize the vector \(\mathbf{x}\) to a \(c\)-bit binary vector is defined as:
\begin{equation}
h(\mathbf{x}) = \operatorname{sign} \left[ \left( \boldsymbol{\kappa}(\mathbf{x}) - \mu \cdot \mathbf{1} \right)^\top A \right]
\end{equation}
where \(\kappa\left(\mathbf{x}\right)=\left[\kappa\left(\mathbf{x}, \mathbf{s}_1\right), \cdots, \kappa\left(\mathbf{x}, \mathbf{s}_m\right)\right]^\top\in\mathbb{R}^m\) is a vector consisting of RBF kernel values between the vector \(\mathbf{x}\) and the support samples \(S\).
\(\mu = \frac{1}{m}\sum_{i=1}^{m}\kappa\left(\mathbf{x}, \mathbf{s}_i\right)\) is the mean kernel value, used as a zero-centering normalizer, and \(\mathbf{1}\in\mathbb{R}^m\) a vector of all ones.


For pattern analysis, we adopt single‑head self‑attention instead of MHSA to simplify attention calculation. This is because SLP-based compressed hash embeddings provide sufficient information. Specifically, this contributes to optimization by modularizing SLP-based feature processing and single-head attention-based pattern analysis.


Formally, in the previous section, we obtain representations for \(L\) bins as \(\mathrm{SLP}(X) = [\mathbf{z}_1, \cdots, \mathbf{z}_L]\). 
Using these bin representation,
we compute the query \(\mathbf{q}_i = \mathbf{z}_i W_Q\), key \( \mathbf{k}_i = \mathbf{z}_i W_K\), and value \(\mathbf{v}_i = \mathbf{z}_i W_V \) for each position \(i\), where \(W_Q, W_K, W_V \in \mathbb{R}^{D \times D}\) is a learnable parameter.
Then the attention for query \(\mathbf{q}_j\), based on keys \(K = \{\mathbf{k}_i\}\) and values \(V = \{\mathbf{v}_i\}\), is defined as:
\begin{equation}
\operatorname{Attn}(\mathbf{q}_j, K, V)=\frac{h(\mathbf{q}_j)^{\top}\mathbf{M}(K, V)}{h(\mathbf{q}_j)^{\top}\overline{\mathbf{k}}}
\end{equation}
where \(\mathbf{M}(K, V) = \sum_{i=1}^{L} h(\mathbf{k}_i) \otimes \mathbf{v}_i\) denotes the key–value memory map, and \(\overline{\mathbf{k}} = \sum_{i=1}^{L} h(\mathbf{k}_i)\) is the sum of the key vector.
We omit the bias term in attention for clarity.

Note that this formulation complements the SLP mechanism, reducing attention complexity to \(\mathcal{O}(\log N)\). 
Furthermore, since both the inner and outer products are reduced to signed additions over binary vectors.
For example, multiplying by \(1\) preserves the original value, while multiplying by \(-1\) simply flips its sign.
Moreover, the memory map \(\mathbf{M}(K, V)\) and the key sum \(\overline{\mathbf{k}}\) can be precomputed and reused without modification.
As a result, \textsc{Scare} improves both computational efficiency and HW compatibility.

Finally, we employ a standard feed-forward network (FFN). While recent SOTA approach \cite{ahn2025real} has demonstrated performance using a single linear layer as a FFN, we leverage the FFN to reconstruct meaningful representations from the binarized computations. Although this incurs additional overhead, it remains efficient due to the significant cost reductions achieved through token compression and hash attention.



\subsection{Hash Function Optimization}
\label{sec:opt}

Optimization of the binary hash function $h$ presents two major challenges. First, the discrete nature of the sign function causes gradient propagation problems, which we address with STE. Second, the exponential growth of possible hash code combinations makes training unstable. To mitigate this problem, existing research has introduced auxiliary learning such as Hamming affinity \cite{liu2022ecoformer}, but this increases training costs and hinders end‑to‑end learning.

In this study, we utilize the reduced token count by SLP as a key solution. The compressed tokens significantly reduce the search space of hash code combinations, allowing sufficient binarization performance to be achieved with RBF kernel‑based similarity approximation alone. Therefore, effective learning is possible with the objective function alone without complex auxiliary learning. Nevertheless, optimization using the full dataset still requires considerable training time and computational cost. 

To reduce this, we propose a dynamic support set sampling strategy that randomly extracts a small number of samples from each mini‑batch during training.
Specifically, from lens representation \(Z^{(b)}=\mathrm{SLP}(X^{(b)})\) in each mini‑batch \(b\), we uniformly extract \(m\) samples to construct \(S^{(b)} = \{s_{1}^{(b)}, s_{2}^{(b)}, \dots, s_{m}^{(b)}\}\). 
Notably, this significantly reduces computational overhead while reflecting diverse query‑key relationships, and achieves robust performance without overfitting to specific patterns. During inference, we use a fixed set to improve efficiency.
We omit the batch index \(b\) in Section \ref{sec:LHA} for simplicity, because all batches follow the same procedure.

Consequently, this study effectively solves the hash function optimization problem by integrating SLP-based compressed token structure with STE, RBF kernel, and support set sampling strategy. It supports end-to-end learning without separate auxiliary learning, and ensures stability and reliability in extreme situations such as sudden input changes, enabling efficient deployment in practical sensor calibration.

\begin{table*}[t]
\centering
\footnotesize
\setlength\tabcolsep{4.5pt}
\renewcommand{\arraystretch}{1.15}

\resizebox{\textwidth}{!}{%
\begin{tabular}{cccccccccccc}
\toprule
& & \multicolumn{4}{c}{\textbf{Accuracy}}
& \multicolumn{3}{c}{\textbf{Real-time}}
& \multicolumn{3}{c}{\textbf{Resource Efficiency}} \\
\cmidrule(lr){3-6}\cmidrule(lr){7-9}\cmidrule(lr){10-12}

& & \multicolumn{3}{c}{\textbf{Mean}} & \textbf{Instant}
& \textbf{Mean} & \textbf{Max} & \textbf{Sampling}
& \textbf{Complexity} & \textbf{Memory} & \textbf{Compatibility} \\
\cmidrule(lr){3-5}\cmidrule(lr){6-6}\cmidrule(lr){7-9}\cmidrule(lr){10-12}

& & \multicolumn{3}{c}{\makecell[c]{RMSE \scriptsize$\downarrow$ \\(\(\mu\)g m\(^{-3}\))}}
& \multirow{2}{*}{\makecell[c]{Top-5\%\\RMSE \scriptsize$\downarrow$\\(\(\mu\)g m\(^{-3}\))}}
& \multirow{2}{*}{\makecell[c]{Mean\\Latency \scriptsize$\downarrow$\\(ms)}}
& \multirow{2}{*}{\makecell[c]{Max\\Latency \scriptsize$\downarrow$\\(ms)}}
& \multirow{2}{*}{\makecell[c]{Latency\\Jitter \scriptsize$\downarrow$ \\\(\sigma\) (ms)}}
& \multirow{2}{*}{\makecell[c]{Computational\\Complexity \scriptsize$\downarrow$\\(Big-O)}}
& \multirow{2}{*}{\makecell[c]{Peak\\Activation \scriptsize$\downarrow$\\(KB)}}
& \multirow{2}{*}{\makecell[c]{CPU Util\\Max \scriptsize$\downarrow$\\(\%)}} \\
\cmidrule(lr){3-5}

\makecell[c]{\textbf{Category}}
& \makecell[c]{\textbf{Models}}
& \makecell[c]{\(\mathrm{PM}_{10}\)} 
& \makecell[c]{\(\mathrm{PM}_{2.5}\)} 
& \makecell[c]{\(\mathrm{PM}_{1}\)}
&  &  &  &  &  &  &  \\

\midrule

\multirow{3}{*}{\makecell[c]{Timeseries\\Forecasting}}
  & DLinear   & 17.81 & 8.41 & 4.49 & 35.23 & \underline{1.31} & \underline{1.44} & {\first{1.83}} & \(\mathbf{O(N)}\) & \underline{28}    & 50.0 \\
  & PatchTST  & 15.23 & 6.29 & \underline{2.71} & 31.34 & 5.60 & 6.63 & 9.75 & \(\mathbf{O((N/P)^2)}\) & 139   & 100.0 \\
  & Mamba     & 16.06 & 6.37 & 3.14  & 32.80 & 269.45 & 272.43 & 15.96 & \(\mathbf{O(N)}\) & 17616 & 14.6 \\
\cmidrule(l{0.5mm}r{0.5mm}){1-2}
\cmidrule(l{0.5mm}r{0.5mm}){3-6}
\cmidrule(l{0.5mm}r{0.5mm}){7-9}
\cmidrule(l{0.5mm}r{0.5mm}){10-12}

\multirow{1}{*}{Hash}
  & Ecoformer & 16.90 & 7.13 & 3.51 & 35.41 & 14.43 & 15.28 & 8.18 & \(\mathbf{O(N)}\) & 1005 & 100.0 \\
\cmidrule(l{0.5mm}r{0.5mm}){1-2}
\cmidrule(l{0.5mm}r{0.5mm}){3-6}
\cmidrule(l{0.5mm}r{0.5mm}){7-9}
\cmidrule(l{0.5mm}r{0.5mm}){10-12}

\multirow{4}{*}{Calibration}
  & NLinear   & 19.33 & 9.96 & 6.13 & 34.36 & {\first{1.28}} & {\first{1.38}} & 3.03 & \(\mathbf{O(N)}\) & {\first{18}}    & 100.0 \\
  & SenDaL    & 15.76 & 6.87 & 3.14 & 38.77 & 32.09 & 118.23 & 18.32 & \(\mathbf{O(N^2)}\) & 9614  & 9.5   \\
  & TESLA     & \underline{14.21} & \underline{6.19} & 2.82 & \underline{27.92} & 2.14 & 2.47 & 4.47 & \(\mathbf{O((logN)^2)}\) & 121   & \underline{8.8}   \\


   & \textsc{Scare}   & {\first{14.03}} & {\first{5.67}} & {\first{2.51}}
  & {\first{27.66}} & 1.63 & 1.72 & \underline{2.11} & \underline{\(\mathbf{O(N\log N)}\)} & 115   & {\first{7.1}} \\

\bottomrule
\end{tabular}}
\caption{Evaluation of category-specific representative models with respect to eight microscopic requirements. Calibration accuracy was computed as the mean RMSE over a 360 length window for each pollutant (PM10, PM2.5, and PM1) using data from three sites (Ant., Oslo, and Zag.). Detailed definitions of the remaining metrics are provided in section \ref{sec:04-evaluation-metrics}. For each metric, the highest performance is marked in {\first{bold}}, and the second highest in \underline{underlined}.}
\label{tb:result_main}
\vspace{0.5em}
\end{table*}

\begin{table*}[t]
\centering
\footnotesize
\setlength\tabcolsep{8pt}
\renewcommand{\arraystretch}{1.15}
\resizebox{\textwidth}{!}{%
\begin{tabular}{c cccc ccc c cc}
\toprule
\multirow{2}{*}{\textbf{Exp}} & \multicolumn{4}{c}{\textbf{Methodology}} & \multicolumn{3}{c}{\textbf{RMSE $\downarrow$}} & \multirow{2.5}{*}{\textbf{\makecell[c]{Max\\Latency $\downarrow$}}} & \multirow{2.5}{*}{\textbf{\makecell[c]{Peak\\Activation $\downarrow$}}} & \multirow{2.5}{*}{\textbf{\makecell[c]{CPU Util\\Max $\downarrow$}}} \\ 
\cmidrule(lr){2-5}\cmidrule(lr){6-8}
& \textbf{Embedding} & \textbf{Patching} & \textbf{Attention} & \textbf{Sampling} & $\mathrm{PM}_{10}$ & $\mathrm{PM}_{2.5}$ & $\mathrm{PM}_{1}$ &  &  &  \\
\midrule
1 & Local         & None           & MHA             & X & 14.87         & \first{5.49}           & 2.59 & 120.32 & 9444.9 & 100.0 \\
2 & Local         & SLP            & MHA             & X & 14.65         & 6.05           & \underline{2.56} & \first{1.69} & \first{109.3}  & 66.7 \\
3 & Local + Global & SLP           & MHA             & X & 14.30         & 5.90           & 2.82 & 3.01  & 133.3  & 75.0 \\
4 & Local + Global & SLP           & HA  & X & \underline{14.13}         & 5.93           & 3.11 & 4.23  & 124.2  & 13.7 \\
5 & Local + Global & SLP           & EBA             & X & 14.18          & 5.97         & 2.90 & 3.15  & 118    & \underline{12.4} \\
\;\,6${}^*$ &Local + Global & SLP           & EBA             & O & \first{14.03}         & \underline{5.67}           & \first{2.51} & \underline{1.72} & \underline{115.0} & \first{7.1} \\
\bottomrule
\end{tabular}}
\caption{Comparison of architectural combinations in terms of mean accuracy, maximum inference latency, memory, and HW compatibility when the input window length is 360. Mean accuracy is calculated as RMSE averaged across the three regions (Ant., Oslo, Zag.) for each particulate matter category. Exp 6${}^*$ is our proposed model. In Attention part, MHA denotes multi-head attention, HA denotes conventional hash attention \cite{liu2022ecoformer}, and EBA denotes efficient bitwise attention.}
\label{tb:architecture_comparison}
\vspace{0.5em}
\end{table*}

\section{Experiment}
\label{sec:04-experiment}
We introduce the datasets and training setup, then outline baselines, evaluation metrics, and implementation details.

\paragraph{Evaluation setup. \label{sec:04-experiment-setup}} Following the experimental protocol of \cite{ahn2025real}, our experiments use a large-scale calibration dataset comprising sensor readings from three cities—Antwerp (\textbf{Ant.}), Oslo (\textbf{Oslo}), and Zagreb (\textbf{Zag.})—each recording three particulate-matter measures (\(\textbf{PM}_\textbf{10}\), \(\textbf{PM}_\textbf{2.5}\), and \(\textbf{PM}_\textbf{1}\)). We assume each region contains multiple sensors of the same type, each uniquely named. Within each region and for each feature, we sort the sensors alphabetically, assign the second‑to‑last sensor to validation, the last to testing, and use all others for training.

\paragraph{Baselines. \label{sec:04-baseline}} 
To compare the performance of \textsc{Scare}, we carefully select commonly used state-of-the-art (SOTA) models across various tasks in IoT systems. For time-series forecasting, we include DLinear \cite{zeng2023dlinear}, PatchTST \cite{nie2023timeseriesworth64}, and Mamba \cite{dao2024transformers_ssm}, which have demonstrated strong predictive capabilities but have not been thoroughly evaluated for sensor calibration. For hash-based approach, we adopt EcoFormer \cite{liu2022ecoformer}, which leverages trainable hash attention to generate compact query–key codes. For sensor calibration, we compare NLinear \cite{zeng2023dlinear}, SenDaL \cite{ahn24sendal}, and TESLA \cite{ahn2025real}, with TESLA achieving SOTA results in this task (Detailed explanations for each baseline in the supplementary material).



\paragraph{Evaluation metrics. \label{sec:04-evaluation-metrics}}
Performance is evaluated as follows. Accuracy is measured by RMSE to penalize large deviations. Responsiveness to sudden changes is assessed via RMSE over the top 5\% of time points with largest absolute first difference \( |\Delta x| \). Real-time behavior is measured by 50 repeated inferences to obtain mean, max, and jitter (sample standard deviation) of latency, capturing throughput, deadline safety, and timing variation. Computational cost is expressed in \( \mathcal{O}(\cdot) \) to show how inference time scales with input length independent of HW. Memory demand is estimated as the peak activation size (sum of retained tensor elements $\times$ bytes per element) in kilobytes. HW suitability is measured by the maximum CPU utilization in repeated inference to indicate typical load and potential short-term spikes.


\paragraph{Implementation. \label{sec:04-implementation}}
The on-device environment was CPU-based (AMD EPYC 7513). Models were developed with TensorFlow 2.14 and trained for 10 epochs using 32 batch size, mean squared error (MSE) loss, and Adam optimizer, following common setups \cite{liu2024itransformer, ahn2025real}. Trained models were converted to TensorFlow Lite FlatBuffers and deployed on an Arduino Nano 33 BLE Sense.

\section{Experimental Results}
This section details our experimental results.
\paragraph{Microscopic requirement.} 
Table \ref{tb:result_main} demonstrates that \textsc{Scare} outperforms existing baseline models in meeting all microscopic requirements. It achieves the highest performance across pollutant types even under rapid and temporary distribution shifts. While linear models such as DLinear and NLinear typically enable faster inference, \textsc{Scare} offers near-linear response times with substantial accuracy gains. In terms of resource efficiency, it exhibits the lowest computational complexity among nonlinear models and surpasses all benchmarks in both maximum CPU utilization and minimum runtime memory requirements.


\paragraph{Ablation study.} 
We introduced six architectural combinations and evaluated their impacts, as summarized in Table \ref{tb:architecture_comparison}. First, switching from simple tokenization (Exp 1) to the SLP patching method (Exp 2) shown better results across all metrics, including accuracy. Second, integrating local and global embeddings (Exp 3) improved mean accuracy compared to using only local embeddings (Exp 2), indicating these embeddings complement each other effectively. Third, substituting conventional hash attention (Exp 4) with proposed EBA (Exp 5 and 6) mitigated information loss while significantly reducing maximum inference latency, peak CPU utilization, and minimum runtime memory requirements.



\paragraph{Model adaptability.} 
Table \ref{fig:interval_board} evaluates adaptability of \textsc{Scare} to sudden input changes. Segments with error deviations from prior inputs were sampled with predefined thresholds. Typically, Ant contributes to model inaccuracies with small error deviations, while Zag complicates calibration with segments with large error deviations. Specifically, DLinear and EcoFormer demonstrate that conventional linear and nonlinear approaches struggle with the calibration task, and the SOTA model TESLA shows limited performance on Ant. Conversely, \textsc{Scare} consistently offers robust performance across all cities and thresholds, with its advantage over other models becoming increasingly evident at various deviation rates.



\paragraph{On-device performance.} 
To assess on-device performance in IoT environments, we evaluated the models on a MCU for deployment and inference latency, as summarized in Figure \ref{fig:exp_board}. DLinear is compact and capable of handling long sequences but provides poor accuracy. TESLA showed high accuracy with inference speeds close to DLinear, but it failed to operate on ultra-long time-series inputs (1440 sequence length). Conversely, \textsc{Scare} not only achieved SOTA accuracy but also maintained latency similar to DLinear across all sequence lengths, while achieving a smaller Flatbuffer size than TESLA.

\section{Discussions and Limitations}

We discuss the implications and challenges of our findings.

\begin{figure}[t]
  \centering
  \includegraphics[
    width=\linewidth,      
    trim=10 0 10 0,        
    clip                  
  ]{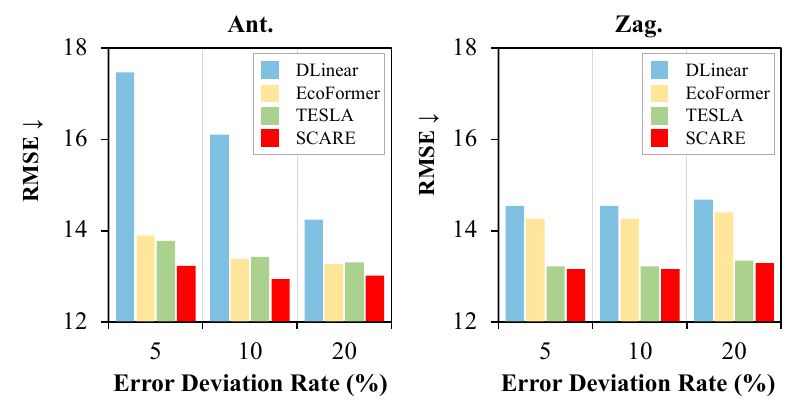}
  \vspace{-1.5em}
  \caption{RMSE comparisons within ±30 time points around change points, under error change thresholds of 5\%, 10\%, 20\% on the Ant. and Zag. datasets.}
  \label{fig:interval_board}
\end{figure}

\paragraph{Robust accuracy.}
As shown in Table \ref{tb:architecture_comparison}, shifting from simple tokenization (Exp 1) to the SLP operation method (Exp 2) led to a statistically significant increase in mean accuracy. This improvement indicates that \textsc{Scare} more effectively preserves salient patterns within each bin, resulting in overall performance gains. Moreover, our dynamic sampling strategy significantly enhances model adaptability, as evidenced by substantial improvements of instant accuracy shown in Figure \ref{fig:interval_board} and Table \ref{tb:architecture_comparison}. These results demonstrate that \textsc{Scare} not only consistently outperforms both linear and nonlinear baselines but also ensures robustness under sudden distribution shifts.

\paragraph{Real-time performance.}
\textsc{Scare} achieves the lowest mean and max latency among all nonlinear models in Table \ref{tb:result_main}. This efficiency stems from reduced token count processed by the SLP stage, which significantly decreases bitwise attention operation during the EBA stage. Moreover, \textsc{Scare} exhibits minimal sampling jitter in Table \ref{tb:result_main}, indicating stable and consistent calibration cycles.

MCU environments require optimized architecture due to the resource constraints such as limited parallelism. While Mamba achieves linear complexity, it requires parallel operations that impose high RAM usage on CPU. Conversely, \textsc{Scare} employs single-head attention to eliminate parallel operation, ensuring fast inference on MCUs without additional memory overhead. 

\begin{figure}[t]
  \centering
  \vspace{0.9em}
  \includegraphics[
    width=\linewidth,      
    trim=5 0 5 0,        
    clip                   
  ]{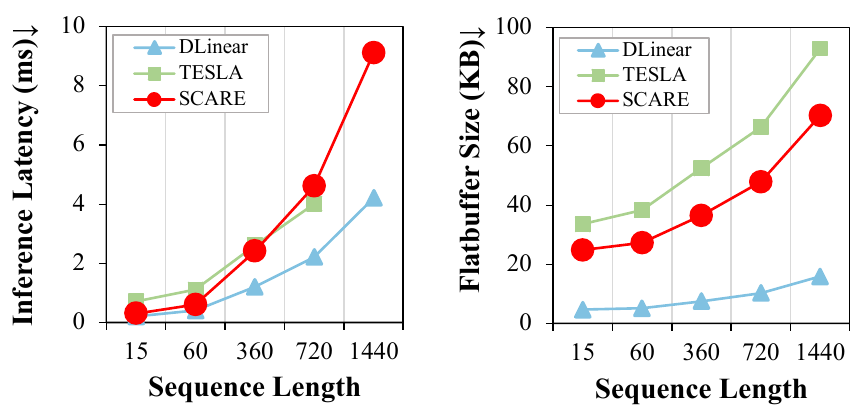}
  \caption{Performance on Arduino Nano 33 BLE Sense. The graphs show inference latency and flat-buffer size across varying windows (15, 60, 360, 720, 1440). Missing points indicate that on-device measurement was not possible.}
  \label{fig:exp_board}
\end{figure}

\paragraph{HW compatibility.}
In Table \ref{tb:result_main}, the CPU utilization results show that \textsc{Scare} achieves the lowest CPU load and memory usage during inference. This efficiency results from the synergy between SLP operations and EBA mechanisms, which maintain linear computational complexity while minimizing overall computation. The SLP's reliance on a single weight matrix $W$ and global bias for uniform bin handling integrates seamlessly with DMA and vector processing units, enabling highly efficient MCU execution. Table \ref{tb:architecture_comparison} confirms that applying SLP simultaneously reduces peak activated memory and CPU utilization when comparing Exp 1 (vanilla transformer) and Exp 2 (SLP). Moreover, Figure \ref{fig:exp_board} demonstrates that while TESLA models fail to execute ultra-long sequences on Arduino due to memory constraints despite robust calibration performance on standard sequences, \textsc{Scare} reliably handles extended data streams through its simplified EBA layer, confirming strong hardware compatibility for real-time embedded deployments.

\paragraph{Architectural trade-offs.}
\textsc{Scare} is deliberately engineered as an ultra-compressed architecture optimized for embedded devices, intentionally omitting complex components such as a standard MHSA. While this structural simplification enables real-time processing in resource-constrained environments (e.g., MCUs), it imposes limitations on representational capacity and flexibility compared to elaborate models designed for high-end platforms. This design prioritizes real-time responsiveness and stability in resource-scarce settings, allowing \textsc{Scare} to deliver consistent, efficient performance on low-specification hardware. In contexts with abundant computational resources, model complexity could be increased to pursue further accuracy gains; however, this study focuses on ensuring reliability in edge environments.

\paragraph{Learning limitations.}
\textsc{Scare} relies on offline retraining and static TensorFlow Lite deployments, preventing devices from correcting seasonal or long-term sensor drift until the next firmware update. This limitation increases operation and maintenance costs and risks service outages in areas with poor connectivity. Future work will incorporate lightweight on-device continual learning through fine-tuning SLP-compressed parameters at low precision, along with server-edge federated calibration to enable real-time adaptation to sensor drift.


\section{Conclusion}
\label{sec:07-conclusion}
In this work, we introduce SCARE, an efficient on-device sensor calibration transformer for low-cost sensors. We first recognize that the conventional accuracy, real-time, resource efficiency triad masks real-world bottlenecks, and therefore decomposed it into eight microscopic requirements to better reflect the realities of on-device operation. SCARE meets eight core requirements through three core components: (i) SLP that effectively compresses time-series data while preserving cross-bin patterns, reducing attention complexity to O(N log N); (ii) EBA that reduces computational and memory complexity to near-linear scale through hash-attention mechanism with single-head attention; and (iii) hash function optimization, which a dynamic sampling strategy ensure both MCU compatibility and millisecond-level inference with robust performance. Evaluations on MCU units deployments show that \textsc{Scare} satisfies all eight microscopic requirements while simultaneously outperforming state-of-the-art linear, hybrid, and deep-learning calibrators in accuracy, real-time, and resource efficiency.


\appendix

\bibliography{aaai2026}

@article{koziel2025efficient,
  title={Efficient field correction of low-cost particulate matter sensors using machine learning, mixed multiplicative/additive scaling and extended calibration inputs},
  author={Koziel, Slawomir and Pietrenko-Dabrowska, Anna and Wojcikowski, Marek and Pankiewicz, Bogdan},
  journal={Scientific Reports},
  volume={15},
  number={1},
  pages={18573},
  year={2025},
  publisher={Nature Publishing Group UK London}
}

@inproceedings{yu2020airnet,
  title={Airnet: A calibration model for low-cost air monitoring sensors using dual sequence encoder networks},
  author={Yu, Haomin and Li, Qingyong and Geng, Yangli-ao and Zhang, Yingjun and Wei, Zhi},
  booktitle={Proceedings of the AAAI conference on artificial intelligence},
  volume={34},
  number={01},
  pages={1129--1136},
  year={2020}
}

@inproceedings{dao2024transformers_ssm,
  title     = {Transformers are SSMs: Generalized Models and Efficient Algorithms Through Structured State Space Duality},
  author    = {Dao, Tri and Gu, Albert},
  booktitle = {Proceedings of the 41st International Conference on Machine Learning},
  series    = {Proceedings of Machine Learning Research},
  volume    = {235},
  year      = {2024},
  pages     = {1--13},
  address   = {Vienna, Austria},
  publisher = {PMLR},
  url       = {https://proceedings.mlr.press/v235/dao24a.html}
}

@article{liang2023mcuformer,
  title={Mcuformer: Deploying vision tranformers on microcontrollers with limited memory},
  author={Liang, Yinan and Wang, Ziwei and Xu, Xiuwei and Tang, Yansong and Zhou, Jie and Lu, Jiwen},
  journal={Advances in Neural Information Processing Systems},
  volume={36},
  pages={8501--8512},
  year={2023}
}

@article{lin2020mcunet,
  title={Mcunet: Tiny deep learning on iot devices},
  author={Lin, Ji and Chen, Wei-Ming and Lin, Yujun and Gan, Chuang and Han, Song and others},
  journal={Advances in neural information processing systems},
  volume={33},
  pages={11711--11722},
  year={2020}
}

@article{ni2025energy,
  title={Energy-aware edge computing optimization for real-time anomaly detection in IoT networks},
  author={Ni, Chunhe and Wu, Jiang and Wang, Hongbo},
  journal={Applied and Computational Engineering},
  volume={139},
  pages={42--53},
  year={2025}
}

@article{zhao2025slopt,
  title={SLOpt: Serving real-time inference pipeline with strict latency constraint},
  author={Zhao, Zhixin and Hu, Yitao and Yang, Guotao and Gong, Ziqi and Shen, Chen and Zhao, Laiping and Li, Wenxin and Liu, Xiulong and Qu, Wenyu},
  journal={IEEE Transactions on Computers},
  year={2025},
  publisher={IEEE}
}

@inproceedings{yang2021deeprt,
  title={Deeprt: A soft real time scheduler for computer vision applications on the edge},
  author={Yang, Zhe and Nahrstedt, Klara and Guo, Hongpeng and Zhou, Qian},
  booktitle={2021 IEEE/ACM Symposium on Edge Computing (SEC)},
  pages={271--284},
  year={2021},
  organization={IEEE}
}

@article{rahman2020systematic,
  title={A systematic review on performance evaluation metric selection method for IoT-based applications--ScienceDirect},
  author={Rahman, MA and Rahman, MS and Hossain, MA and Al-Fuqaha, A and Wang, Y},
  journal={Journal of Network and Computer Applications},
  year={2020}
}

@inproceedings{ibrahim2024holistic,
  title={Holistic evaluation metrics for federated learning},
  author={Ibrahim, Jehad and Li, Yanli and Chen, Huaming and Yuan, Dong},
  booktitle={2024 27th International Conference on Computer Supported Cooperative Work in Design (CSCWD)},
  pages={2282--2287},
  year={2024},
  organization={IEEE}
}

@inproceedings{kitaev2020reformer,
  title     = {Reformer: The efficient transformer},
  author    = {Kitaev, Nikita and Kaiser, {\L}ukasz and Levskaya, Anselm},
  booktitle = {Proceedings of the International Conference on Learning Representations},
  year      = {2020},
  url       = {https://openreview.net/forum?id=rkgNKkHtvB}
}

@article{villanueva2023mutlisensor, 
title={Smart Multi-Sensor Calibration of Low-Cost Particulate Matter Monitors}, 
volume={23}, 
ISSN={1424-8220}, 
DOI={10.3390/s23073776}, 
number={7}, 
journal={Sensors}, 
publisher={MDPI AG}, 
author={Villanueva, Edwin and Espezua, Soledad and Castelar, George and Diaz, Kyara and Ingaroca, Erick}, 
year={2023}, 
month={Apr}, 
pages={3776} }

@inproceedings{qiu2025enhancing,
  title={Enhancing Masked Time-Series Modeling via Dropping Patches},
  author={Qiu, Tianyu and Xie, Yi and Niu, Hao and Xiong, Yun and Gao, Xiaofeng},
  booktitle={Proceedings of the AAAI Conference on Artificial Intelligence},
  volume={39},
  number={19},
  pages={20077--20085},
  year={2025}
}

@inproceedings{chen2025sequence,
  title={Sequence complementor: Complementing transformers for time series forecasting with learnable sequences},
  author={Chen, Xiwen and Qiu, Peijie and Zhu, Wenhui and Li, Huayu and Wang, Hao and Sotiras, Aristeidis and Wang, Yalin and Razi, Abolfazl},
  booktitle={Proceedings of the AAAI Conference on Artificial Intelligence},
  volume={39},
  number={15},
  pages={15913--15921},
  year={2025}
}

@inproceedings{nie2023timeseriesworth64,
  title     = {A Time Series is Worth 64 Words: Long-term Forecasting with Transformers},
  author    = {Nie, Yuqi and
               H. Nguyen, Nam and
               Sinthong, Phanwadee and 
               Kalagnanam, Jayant},
  booktitle = {International Conference on Learning Representations},
  year      = {2023}
}

@inproceedings{zeng2023dlinear,
  title={Are transformers effective for time series forecasting?},
  author={Zeng, Ailing and Chen, Muxi and Zhang, Lei and Xu, Qiang},
  booktitle={Proceedings of the AAAI conference on artificial intelligence},
  pages={11121--11128},
  year={2023}
}

@inproceedings{liu2024itransformer,
  title     = {iTransformer: Inverted Transformers Are Effective for Time Series Forecasting},
  author    = {Liu, Yong and Hu, Tengge and Zhang, Haoran and Wu, Haixu and Wang, Shiyu and Ma, Lintao and Long, Mingsheng},
  booktitle = {International Conference on Learning Representations},
  year      = {2024}
}

@inproceedings{wan2020alert,
  title={Accurate learning for energy and timeliness},
  author={Wan, Chengcheng and Santriaji, Muhammad and Rogers, Eri and Hoffmann, Henry and Maire, Michael and Lu, Shan},
  booktitle={2020 USENIX annual technical conference (USENIX ATC 20)},
  pages={353--369},
  year={2020}
}

@inproceedings{ahn2025real,
  title={Real-Time Calibration Model for Low-Cost Sensor in Fine-Grained Time Series},
  author={Ahn, Seokho and Kim, Hyungjin and Shin, Sungbok and Seo, Young-Duk},
  booktitle={Proceedings of the AAAI Conference on Artificial Intelligence},
  volume={39},
  number={1},
  pages={3--11},
  year={2025}
}

@article{liu2022ecoformer,
  title={Ecoformer: Energy-saving attention with linear complexity},
  author={Liu, Jing and Pan, Zizheng and He, Haoyu and Cai, Jianfei and Zhuang, Bohan},
  journal={Advances in Neural Information Processing Systems},
  volume={35},
  pages={10295--10308},
  year={2022}
}

@ARTICLE{ahn24sendal,
  author={Ahn, Seokho and Kim, Hyungjin and Lee, Euijong and Seo, Young-Duk},
  journal={IEEE Internet of Things Journal}, 
  title={SenDaL: An Effective and Efficient Calibration Framework of Low-Cost Sensors for Daily Life}, 
  year={2024},
  volume={11},
  number={11},
  pages={20619-20630},
  doi={10.1109/JIOT.2024.3371150}}

@article{concas2021machinelearningcalibration,
author = {Concas, Francesco and Mineraud, Julien and Lagerspetz, Eemil and Varjonen, Samu and Liu, Xiaoli and Puolam\"{a}ki, Kai and Nurmi, Petteri and Tarkoma, Sasu},
title = {Low-Cost Outdoor Air Quality Monitoring and Sensor Calibration: A Survey and Critical Analysis},
year = {2021},
issue_date = {May 2021},
publisher = {Association for Computing Machinery},
address = {New York, NY, USA},
volume = {17},
number = {2},
issn = {1550-4859},
journal = {ACM Trans. Sen. Netw.},
month = {may},
articleno = {20},
numpages = {44}
}

@inproceedings{rastegari2016xnor,
  title={Xnor-net: Imagenet classification using binary convolutional neural networks},
  author={Rastegari, Mohammad and Ordonez, Vicente and Redmon, Joseph and Farhadi, Ali},
  booktitle={European conference on computer vision},
  pages={525--542},
  year={2016},
  organization={Springer}
}

@inproceedings{liu2018bi,
  title={Bi-real net: Enhancing the performance of 1-bit cnns with improved representational capability and advanced training algorithm},
  author={Liu, Zechun and Wu, Baoyuan and Luo, Wenhan and Yang, Xin and Liu, Wei and Cheng, Kwang-Ting},
  booktitle={Proceedings of the European conference on computer vision (ECCV)},
  pages={722--737},
  year={2018}
}

\end{document}